\documentclass[runningheads]{llncs}
\usepackage[T1]{fontenc}
\usepackage{graphicx}
\usepackage{booktabs}
\usepackage[misc]{ifsym}
\newcommand{\corr}{(\Letter)}
\usepackage{mwe}

\usepackage{times}
\usepackage{soul}
\usepackage{url}
\usepackage[hidelinks]{hyperref}
\usepackage{amsmath}
\usepackage{amssymb}

\usepackage{dsfont}
\usepackage{booktabs}
\usepackage{algorithm}
\usepackage{algorithmic}
\usepackage[switch]{lineno}
\usepackage{wrapfig}

\usepackage{comment}

\usepackage{mathtools}
\usepackage{xcolor}

\usepackage{subcaption}
\usepackage{multirow}
\usepackage{bm}

\newcommand{\ie}{\textit{i}.\textit{e}.,}
\newcommand{\eg}{\textit{e}.\textit{g}.,}
\newcommand{\etc}{\textit{etc}.}
\newcommand{\sota}{state-of-the-art}
\renewcommand{\eqref}[1]{Eq.\,(\ref{#1})}

\newcommand{\romsm}[1]{\lowercase\expandafter{\romannumeral #1\relax}}

\tocauthor{Fudong Lin, Wanrou Du, Jinchan Liu, Tarikul Milon, Shelby Meche, Wu Xu, Xiaoqi Qin, Xu Yuan}
\toctitle{Do Protein Transformers Have Biological Intelligence?}

\begin{document}

\title{Do Protein Transformers Have Biological Intelligence?}


%
\author{Fudong Lin\inst{1} \and
Wanrou Du\inst{2}  \and
Jinchan Liu\inst{3} \and
Tarikul Milon\inst{4} \and
Shelby Meche\inst{4} \and
Wu Xu\inst{4} \and
Xiaoqi Qin\inst{2} \and
Xu Yuan\inst{1} \corr}


\authorrunning{F. Lin, W. Du, et al.}


\institute{University of Delaware, Newark, DE 19716, USA \\
\email{\{fudong,xyuan\}@udel.edu}
\and
Beijing University of Posts and Telecommunications, Haidian, Beijing 100876, China \email{\{wanroudu,xiaoqiqin\}@bupt.edu.cn}
\and
Yale University, New Haven, CT 06520, USA \\
\email{jinchan.liu@yale.edu}
\and University of Louisiana at Lafayette, Lafayette, LA 70504, USA
\email{\{tarikul-islam.milon1,wu.xu\}@louisiana.edu}
}

\maketitle              

\begin{abstract}
    Deep neural networks, particularly Transformers,
    have been widely adopted for predicting the functional properties of proteins. 
    In this work, we focus on exploring whether Protein Transformers 
    can capture biological intelligence among protein sequences.
    To achieve our goal, 
    we first introduce a protein function dataset, namely \textit{Protein-FN}, 
    providing over $9000$ protein data with meaningful labels. 
    Second, we devise a new Transformer architecture,
    namely \textit{Sequence Protein Transformers (SPT)},
    for computationally efficient protein function predictions.
    Third, we develop a novel Explainable Artificial Intelligence (XAI) technique 
    called \textit{Sequence Score}, 
    which can efficiently interpret the decision-making processes of protein models, 
    thereby overcoming the difficulty of deciphering biological intelligence 
    bided in Protein Transformers. 
    Remarkably, even our smallest SPT-Tiny model, 
    which contains only 5.4M parameters, 
    demonstrates impressive predictive accuracy, 
    achieving $94.3 \%$ on the Antibiotic Resistance (AR) dataset 
    and $99.6 \%$ 
    on the Protein-FN dataset, all accomplished by training from scratch.
    Besides, our Sequence Score technique helps reveal that 
    our SPT models can discover several meaningful patterns 
    underlying the sequence structures of protein data, 
    with these patterns aligning closely with the domain knowledge in the biology community.
    We have officially released our Protein-FN dataset on Hugging Face Datasets \textcolor{magenta}{\url{https://huggingface.co/datasets/Protein-FN/Protein-FN}}.
    Our code is available at \textcolor{magenta}{\url{https://github.com/fudong03/BioIntelligence}}.
\end{abstract}

\section{Introduction}
\label{sec:intro}

Proteins serve as the architects of life, orchestrating an extraordinary range of functions that bring vitality and complexity to the biological world.
Their roles encompass everything from catalyzing critical biochemical reactions to facilitating precise cellular communication.
Decoding the intricate relationship between a protein's sequence, structure, and functional properties holds the key to unraveling these life-sustaining mysteries.
This endeavor is more than a scientific pursuit; it is a profound exploration of the fundamental processes that define life itself.

Since the intricate patterns of protein sequences are analogous to 
the syntactic and semantic structures found in human languages, 
existing \sota\ 
Protein Language Models (PLMs)~\cite{rao:nips19:tape,rives:pans21:ESM-1b,meier:nips21:esm_1v,rao:icml21:msa_transformer,jumper:nature21:alpha_fold,evans2021protein,baek2021accurate,luo2022antigen,hsu2022learning,hu2022exploring}
harness the advanced language models~\cite{vaswani:nips17:attn,devlin:naacl_hlt19:bert}
to decipher how the intricate structures of protein sequences dictate their functional properties.
%
However,
these methods require pre-training on millions or even billions of protein sequences for satisfactory performance.
The excessive computational demands of self-supervised pre-training 
render PLMs unattainable for resource-constrained research groups.

In this work, we develop a computationally efficient Transformer architecture, 
namely Sequence Protein Transformer (SPT), 
for unraveling the complex interplay between a protein's sequence and its functional property, 
by leveraging the Transformer architectures in the vision domain~\cite{dosovitskiy:iclr21:vit,touvron:icml21:deit,liu:iccv21:swin-vit,kaiming:cvpr22:mae}.
%
Specifically, 
our work focuses on answering the following research question:
\textit{Can our Protein Transformers learn biological intelligence underlined in protein sequences?}

To help answer this question,
we first introduce a new protein function dataset called \textit{Protein-FN},
offering over $9000$ protein data, 
with each containing the protein's 1D amino acid sequences, 3D structures, 
as well as its functional properties annotated by biological experts of our team.
Second, different from PLMs, where the protein sequence is naturally encoded by the letter abbreviation of amino acids
(\eg\ with letter ``A'' for representing the amino acid ``Alanine''),
how to encode the protein data for new Transformer architecture remains unexplored,
making the applications of existing Vision Transformers (ViT) variants~\cite{dosovitskiy:iclr21:vit,touvron:icml21:deit,liu:iccv21:swin-vit,wang:iccv21:pvt,arnab:iccv21:vivit,fan:iccv21:mvit,caron:iccv21:dino,bao:iclr22:beit,fudong:ecml23:storm,fudong:kdd24:crop_net,kaiming:cvpr22:mae,fudong:iccv23:mmst_vit,fudong:cikm24:mat,tiankuo:iclr25:bonemet} 
for protein function predictions technically infeasible.
We develop the Sequence Protein Transformer (SPT) model to address this issue.
Featuring an innovative embedding mechanism tailored for protein data, 
our SPT model excels in predicting the functional properties of proteins. 
Remarkably, it can achieve a superior prediction performance 
without relying on computationally extensive self-supervised pre-training.
Third, Explainable Artificial Intelligence (XAI) techniques
\cite{selvaraju:iccv17:grad_cam,Fong:ICCV17:perturbation-based-method3,Fong:ICCV19:perturbation-based-method2,Srinivas:nips19:fullgrad,Barkan:cikm21:gam,Jalwana:cvpr21:gradient-based,Englebert:icpr22:gradient-based},
especially those Transformer-specific solutions
\cite{Abnar:acl20:rollout,Voita:acl19:vit-method,caron:iccv21:dino,Chefer:cvpr21:vit-method,Chefer:iccv21:vit-method,Xie:ijcai23:vit-method},
can offer insightful perspectives into the decision-making mechanisms of deep neural networks (DNNs),
making them suitable for deciphering biological intelligence resided in Protein Transformers.
However, current XAI approaches face significant challenges 
when handling protein sequences that vary in the number of amino acids. 
They either fail to accommodate these variances 
or incur substantial computational burdens,
\eg\ $\mathcal{O} (L^{2} \cdot P^{4})$ for Attention Flow~\cite{Abnar:acl20:rollout},
where $L$ and $P$ represent the model depth 
and the protein sequence length, respectively.
Consequently, these methods prove impractical for 
analyzing the biological insight within protein sequences.
In this study, we introduce the Sequence Score, 
a novel gradient-based XAI approach, 
tailored specifically to manage protein sequences of varying amino acid lengths.
It advances existing Transformer-specific XAI solutions 
through its computational efficiency,
which \textit{scales linearly} with protein sequence length.
This advancement facilitates a more efficient and effective interpretation of the biological intelligence bided in Protein Transformers.

\begin{figure*} [!t] 
    \centering
    \includegraphics[width=.95\textwidth]{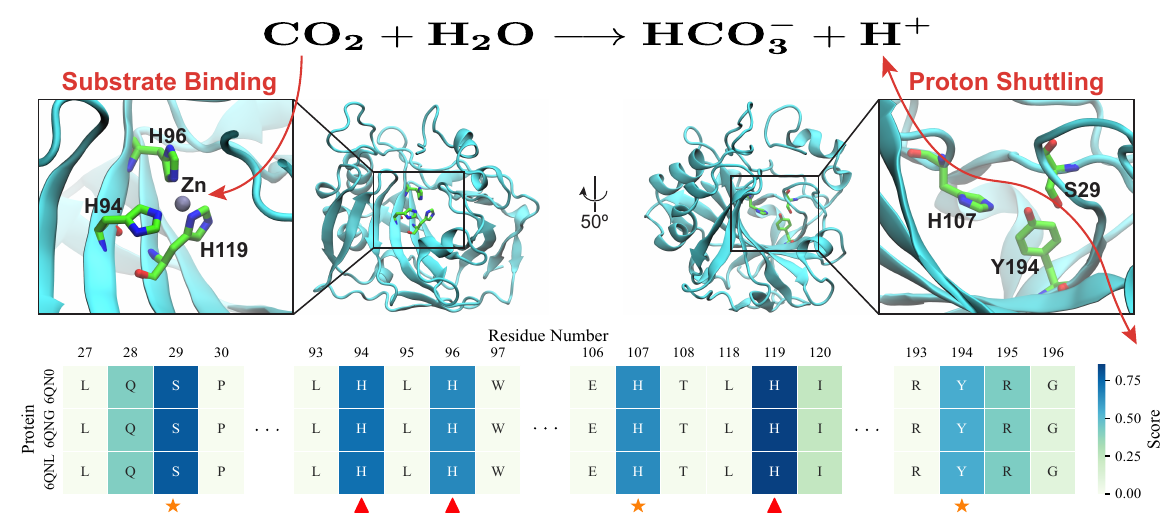}
    \captionof{figure}{Illustration of two key conserved motifs, \ie\ ``His94-His96-His119'' and ``Ser29-His107-Tyr194'',
    for the Carbonic Anhydrases, a vital enzyme class.
    Here, ``His'' (Histidine), ``Ser'' (Serine), and ``Tyr'' (Tyrosine) are the amino acids forming these motifs, 
    abbreviated as ``H'', ``S'', and ``Y'', respectively.
    The residual numbers signify the positions of amino acids within the protein sequence.
    In the upper section, two figures, along with a corresponding equation, 
    utilize the protein ``6QN0'' as an example to 
    visualize the three-dimensional structures of the motifs 
    and elucidate their roles in catalyzing the reaction.
    A heatmap in the lower section displays the importance scores generated by our approach, with the motifs ``His94-His96-His119'' and the ``Ser29-His107-Tyr194'' distinctly marked with red triangles and orange stars, respectively.
    }
    \label{fig:intro-pattern}
\end{figure*}

Our Sequence Score technique has successfully identified several meaningful biological patterns, 
showcasing its prowess in revealing the biological intelligence ingrained in Protein Transformers.
For example, Carbonic Anhydrases (CAs) are a class of enzymes vital to many biological processes, 
such as respiration and acid-base balance in organisms.
Central to the functionality of these enzymes are two highly conserved motifs:
``His94-His96-His119'' and ``Ser29-His107-Tyr194''. 
The ``His94-His96-His119'' pattern, known as the zinc-binding motif, plays a critical role in the catalytic activity of CAs. 
These histidine residues coordinate with a zinc ion, 
which is essential for the hydration of carbon dioxide, 
to realize a primary reaction catalyzed by CAs. 
This interaction is fundamental to 
maintain the enzyme's active site structure and its catalytic efficiency.
On the other hand, the ``Ser29-His107-Tyr194'' motif is crucial for substrate specificity 
and the orientation of water molecules in the active site. 
This motif contributes to the positioning and polarization of water molecules, 
facilitating the transfer of protons and hence supporting the enzyme's catalytic mechanism.
By applying our Sequence Score technique to interpret the prediction results 
of our SPT models,
we discover that our models can capture 
the importance of the ``His94-His96-His119'' and the ``Ser29-His107-Tyr194'' motifs
for the functional properties of CAs.
Figure~\ref{fig:intro-pattern} (see its heatmap) shows the importance scores 
for the three proteins of the CA class,
where our Sequence Score technique assigns very high scores on both
the ``His94-His96-His119'' (marked by red triangles) and 
the ``Ser29-His107-Tyr194'' (marked by orange stars) patterns.
This exhibits that our SPT models have captured biological intelligence 
underlined in the protein sequence for predicting the functional properties of proteins.

\section{Related Work}
\label{sec:rw}

\par\smallskip\noindent
{\bf Protein Language Models.}
Protein Language Models (PLMs) have demonstrated remarkable performance 
across a spectrum of biological tasks.
AlphaFold~\cite{jumper:nature21:alpha_fold} 
is a well-known study in PLMs,
and it uses multiple sequence alignment to predict the 3D structure of proteins.
Recently, PLMs are also developed for predicting the functional properties of proteins,
including single sequence-based methods, 
\ie\ TAPE~\cite{rao:nips19:tape}, ESM-1b~\cite{rives:pans21:ESM-1b},  and ESM-1v~\cite{meier:nips21:esm_1v},
multiple sequence alignment-based approaches,
\ie\ MSA Transformer~\cite{rao:icml21:msa_transformer},
and others~\cite{finn2014pfam,riesselman2018deep,alley2019unified,elnaggar2021prottrans,baek2021accurate,rao:iclr21:transformer,evans2021protein,luo2022antigen,hsu2022learning,hu2022exploring}.
Despite their effectiveness,
PLMs require extensive pre-training on millions of protein data for satisfactory performance,
making them computationally inaccessible for resource-limited groups.
Different from prior studies, 
our SPT model can achieve superb performance in protein function predictions by training from scratch.
Therefore, it advances previous PLMs by significantly reducing the computational overhead.
We hope that the exceptionally computational efficiency of our SPT model 
can shed light on future work in adopting its model architecture 
for protein-relevant tasks.

\par\smallskip\noindent
{\bf XAI Techniques.}
Explainable Artificial Intelligence (XAI) methods
provide valuable insights into the decision-making processes 
of deep neural networks (DNNs),
making them well-suited for interpreting biological intelligence 
underlined in Protein Transformers.
The mainstream XAI techniques targeting DNNs can be roughly grouped into two categories,
\ie\ XAI for CNNs and XAI for Transformers.
The former category is popularized by Grad-CAM~\cite{selvaraju:iccv17:grad_cam},
which weights the activation maps by global-average-pooled gradients flowing into the last convolutional layer.
Subsequently, saliency-based \cite{Dabkowski:nips17:saliency-based-method1,Mahendran:ijcv16:saliency-based-method2,Simonyan:ICLRW14:saliency-based-method3},
activation-based \cite{Zhou:CVPR16:activation-based-method1,Kim:ICML18:activation-based-method2}, 
perturbation-based \cite{Fong:ICCV17:perturbation-based-method3,Fong:ICCV19:perturbation-based-method2,Lundberg:nips17:perturbation-based-method5,petsiuk:bmvc18:deletion,Zeiler:ECCV14:perturbation-based-method1}, 
and gradient-based \cite{Srinivas:nips19:fullgrad,Barkan:cikm21:gam,Sundararajan:icml17:axiom-attr,Jalwana:cvpr21:gradient-based,Englebert:icpr22:gradient-based,Chattopadhyay:wacv18:gradcam++,Fu:bmvc20:axiom-gradcam,Jiang:tip21:layercam,Wang:cvprw20:scorecam,Desai:wacv20:ablationcam,chen2023dark} 
XAI techniques are developed for deciphering the decision-making processes of CNNs.
Despite their popularity,
these methods face challenges when applied to Protein Transformers 
due to the structural differences between Transformers~\cite{vaswani:nips17:attn} and CNNs.
Recently, Attention Rollout and Attention Flow~\cite{Abnar:acl20:rollout},
which map information flow using a Directed Acyclic Graph,
has been proposed to interpret the decision-making processes in Transformer architectures,
with its success inspiring a volume of 
Transformer-specific XAI techniques~\cite{chen2022grease,Voita:acl19:vit-method,Xie:ijcai23:vit-method,caron:iccv21:dino,Chefer:cvpr21:vit-method,Chefer:iccv21:vit-method,chen2022explain}.
However, existing XAI solutions for Protein Transformers 
typically involve substantial computational demands,
\eg\ $\mathcal{O}(L^{2} \cdot P^{4})$ for Attention Flow, 
with $L$ and $P$ respectively representing the depth of the model
and the length of the protein sequence,
making them infeasible to decipher the biological insight
underlined in long protein sequences.
This stems from their requirement to aggregate information from attention weights 
throughout every layer of the Transformer Encoders.
In sharp contrast, 
our Sequence Score technique, while classified in the second category, 
revolutionizes the interpretation of decision-making processes in Transformers.
This achievement stems from its linear complexity 
with respect to the protein sequence length.
Therefore, our solution significantly advances 
previous Transformer-specific XAI techniques in computational efficiency,
making it well-suited for interpreting the biological intelligence resided in Protein Transformers.

\section{The Protein-FN Dataset} 
\label{sec:dataset}

We introduce our curated protein function dataset,
namely \textit{Protein-FN}, 
designed specifically for such biological tasks as 
protein function prediction~\cite{rives:pans21:ESM-1b,meier:nips21:esm_1v}, 
motif identification and discovery~\cite{kondra2021development}, \etc\
Table~\ref{tab:dataset-overview} presents the details of our ProFunc-9K dataset.
This dataset, sourced from the Protein Data Bank (PDB)~\cite{pdb}, 
provides diverse 1D amino acid sequences, 3D protein structures, 
functional properties of $9014$ proteins
($7211$ and $1803$ samples for the training and the test datasets, respectively).
These proteins, after carefully examined by biological experts in our team,  fall into six categories,
\ie\ protease, kinase, receptor, carbonic anhydrase, phosphatase, and isomerase.
Notably, kinases, phosphatases, proteases, and receptors play essential roles in signal
transduction. Most drugs act on proteins involved in signal transduction. Isomerases and carbonic anhydrases are two enzymes that are not directly involved in signal transduction pathways,
but they catalyze critical reactions.

%
%
\begin{table*}   [!t] 
    \scriptsize
    \centering
    \setlength\tabcolsep{5 pt}
    \caption{
        Overview of our Protein-FN dataset
        }
        \begin{tabular}{@{}cccccccc@{}}
            \toprule
            \multirow{2}{*}{Datasets} & \multirow{2}{*}{Samples} & \multicolumn{6}{c}{Classes}                                                                                                                                                 \\ \cmidrule(l){3-8} 
                                      &                            & \multicolumn{1}{c}{Protease} & \multicolumn{1}{c}{Kinase} & \multicolumn{1}{c}{Receptor} & Carbonic Anhydrase & \multicolumn{1}{c}{Phosphatase} & Isomerase \\ \midrule
            Training              & 7211                       & \multicolumn{1}{c}{2439}     & \multicolumn{1}{c}{2003}   & \multicolumn{1}{c}{1172}     & \multicolumn{1}{c}{972}                & \multicolumn{1}{c}{343}         & 282       \\ 
            Test                  & 1803                       & \multicolumn{1}{c}{628}      & \multicolumn{1}{c}{499}    & \multicolumn{1}{c}{265}      & \multicolumn{1}{c}{234}                & \multicolumn{1}{c}{89}          & 88        \\ \midrule
            Total                     & 9014                       & \multicolumn{1}{c}{3067}     & \multicolumn{1}{c}{2502}   & \multicolumn{1}{c}{1437}     & \multicolumn{1}{c}{1206}               & \multicolumn{1}{c}{431}         & 371       \\ \bottomrule
        \end{tabular}
    \label{tab:dataset-overview}
\end{table*}

\section{Our Approaches} 
\label{sec:our_approach}

To unveil the biological intelligence embedded within Protein Transformers, 
we have developed two key innovations: \romsm{1}) the Sequence Protein Transformer (SPT), 
designed for the efficient and effective prediction of protein functions, 
and \romsm{2}) the Sequence Score, 
aimed at efficiently interpreting the decision-making processes of Protein Transformers.

\subsection{Problem Statement}
\label{sec:ps}

Given a protein dataset consisting of $N$ samples, 
denoted as $\mathbb{X} = \{ (\bm{x_{i}}, y_{i}) ~|~ i \in {1, 2, \cdots, N } \}$, 
each sample $\bm{x} \in \mathbb{R}^{P \times 1}$ represents the primary structure (\ie\ the sequence of amino acids) of the protein.
Here, $P$ denotes the sequence length,
and $y \in [C]$ indicates the specific function of the protein,
\eg\ protease, kinase, receptor, \etc\
Notably, the length of the primary structure of proteins, 
\ie\ the number of amino acids in a polypeptide chain, can vary widely in the real scenario.
In this work, our goals are twofold.
First, we aim to develop a simple, computation-efficient Protein Transformer (PT)
$f_{\bm{\theta}}: \mathbb{R}^{P \times 1} \rightarrow [C]$ 
for accurate protein function predictions, 
where $\bm{\theta}$ denotes the PT's hyperparameters.
Second, we plan to propose a novel Explainable Artificial Intelligence (XAI) technique,
which can efficiently interpret long protein sequences.
As such, given a well-trained PT model $f_{\bm{\theta}}$,
 the proposed XAI technique $g: \mathbb{R}^{P \times 1} \rightarrow \mathbb{R}^{P \times 1}$ 
can decode its biological intelligence
by deciphering its decision-making process.

\begin{figure*} [!t] 
    \centering
    \centering
    \includegraphics[width=.98\textwidth]{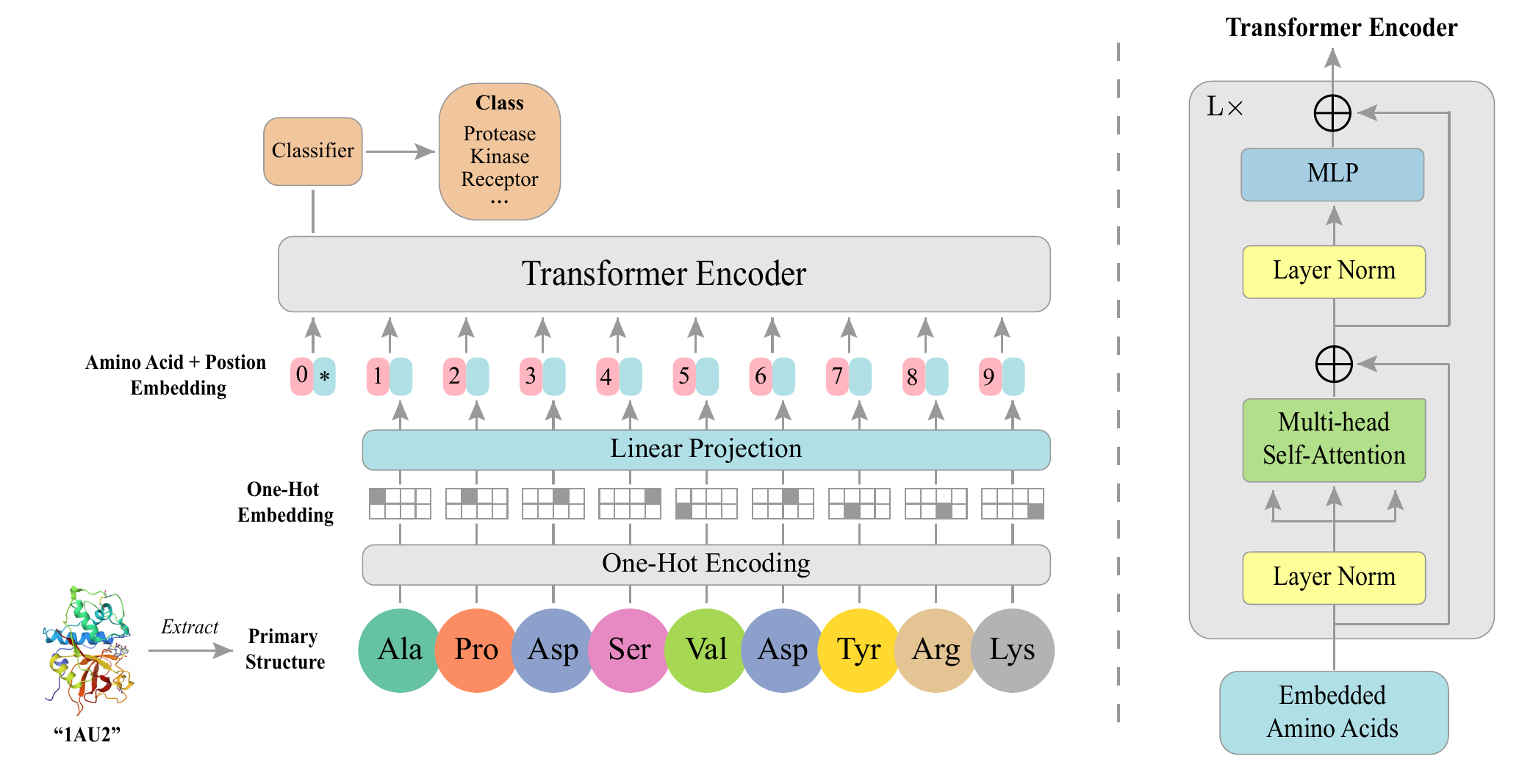}
    \caption{
        The architecture of our Sequence Protein Transformers (SPT) model.
    }
    \label{fig:method-arch-seq-vit}
\end{figure*}

\subsection{Our Sequence Protein Transformer} 
\label{sec:seq-vit}

To achieve our goal,
we propose a simple, computation-efficient Transformer architecture,
namely Sequence Protein Transformer (SPT), 
for predicting the functional properties of proteins.
It can achieve superb prediction performance 
without relying on computation-intensive self-supervised pre-training.

Figure~\ref{fig:method-arch-seq-vit} shows an overview of our model architecture.
We start by extracting the primary structure $\bm{x} \in \mathbb{R}^{P \times 1}$,
\ie\ a sequence of amino acids, from a protein (\eg\ ``1AU2'').
%
%
Then, the one-hot encoding is utilized to encode the sequence of amino acids.
As such, each amino acid is represented by a binary vector of length $d$.
Note that we set $d = 20$ in this study as there are $20$ types of amino acids. 
Next, a linear projection layer $\bm{\text{Proj}}$: $\mathbb{R}^{P \times d} \rightarrow \mathbb{R}^{P \times D}$
is used to project the low dimentional one-hot embedding $\bm{E}_{\textrm{oh}} \in \mathbb{R}^{P \times d}$ 
to the high dimentional amino acid embedding $\bm{E}_{\textrm{ami}} \in \mathbb{R}^{P \times D}$, \ie\
\begin{equation} \label{eq:embed}
    \begin{gathered}
        \bm{E}_{\text{ami}} = \bm{\textrm{Proj}} \left(\textrm{OH} (\bm{x}) \right).
    \end{gathered}
\end{equation}
Here, $D$ is the hidden size of the Transformer Encoder,
and $\textrm{OH}$ represents one-hot encoding.
Similar to prior Transformer variants~\cite{dosovitskiy:iclr21:vit,touvron:icml21:deit,liu:iccv21:swin-vit},
our model prepends a learnable classification token $\bm{E}_{\textrm{cls}} \in \mathbb{R}^{1 \times D}$ 
to the embedding sequence.
As such, the input sequence of the Transformer Encoder can be obtained 
by summing up the amino acid embedding
and the positional embedding $\bm{E}_{\textrm{pos}} \in \mathbb{R}^{(P+1) \times D}$.
Finally, the head of the output sequence $\bm{z} \in \mathbb{R}^{1 \times D}$, encoded by a stack of Transformer blocks, 
is fed to a linear classifier for protein function predictions,
\ie\ $\hat{y} = \bm{W}^{T} \bm{z} + \bm{b}$, 
with $\bm{W}$ and $\bm{b}$ respectively representing the weights and the bias of the classifier, 
and $\hat{y} \in [C]$ indicating the predicted protein function.

The rightmost chart of Figure~\ref{fig:method-arch-seq-vit} depicts 
the architecture of the Transformer Encoder,
where each Transformer block consists of a Multi-head Self-Attention (MSA) block~\cite{vaswani:nips17:attn} and an MLP block.
Each of the two blocks includes a Layer Normalization~\cite{hinton:2016:layer-norm} before the block and the residual connection~\cite{kaiming:cvpr16:resnet} after the block.
Similar to previous studies~\cite{dosovitskiy:iclr21:vit,liu:iccv21:swin-vit,wang:iccv21:pvt,touvron:icml21:deit},
the MLP block is a two-layer neural network with a GELU non-linearity.
Mathematically, our Transformer Encoder can be expressed as below,
\begin{equation} \label{eq:encoder}
    \begin{gathered}
        \bm{E}_{0} = [\bm{E}_{\textrm{cls}}; \bm{E}_{\textrm{ami}}^{1}; \bm{E}_{\textrm{ami}}^{2}; \cdots; \bm{E}_{\textrm{ami}}^{P}] + \bm{E}_{\textrm{pos}},    \\  
        \bm{E}^{\prime}_{\ell} = \textrm{MSA} \left(\textrm{Norm}(\bm{E}_{\ell -1}) \right) + \bm{E}_{\ell-1}, \\
        \bm{E}_{\ell} = \textrm{MLP} \left(\textrm{Norm} (\bm{E}^{\prime}_{\ell}) \right) + \bm{E}^{\prime}_{\ell}, \\
        \bm{z} = \textrm{Norm} (\bm{E}_{L}^{0}).
    \end{gathered}
\end{equation}
Here, $\ell = 1, 2, \cdots, L$ indicates the $\ell$-th block of Transformer Encoder.
Different from prior Transformer variants, our Transformer Encoder has a flexible number of positional embeddings,
enabling the SPT to address the primary structure of proteins,
whose sequence lengths vary significantly.
Inspired by the Multi-head Self-Attention (MSA) mechnism~\cite{vaswani:nips17:attn},
our MSA block here is devised to capture the global protein representation
by learning the dependency among a sequence of amino acids, \ie\
\begin{equation} \label{eq:msa}
    \begin{gathered}
        \textrm{MSA} (\bm{Q}, \bm{K}, \bm{V}) = \textrm{Concat}(\textrm{head}_{1}, \cdots, \textrm{head}_{h}) \bm{W}^{\textrm{O}}, \\
        \textrm{head}_{i} = \textrm{Softmax} (\bm{Q}_{i} \bm{K}_{i}^{T} / \sqrt{d_{k}}) \bm{V}_{i},  \\
        \textrm{where} ~\bm{Q}_{i} = \bm{Q} \bm{W}^{Q}_{i},  ~\bm{K}_{i} = \bm{K} \bm{W}^{K}_{i}, ~\bm{V}_{i} = \bm{V} \bm{W}^{V}_{i}. \\
    \end{gathered}
\end{equation}
Here, $\bm{\textrm{Concat}}$ indicates the operation of feature concatenation,
$h$ is the number of attention heads, 
and $d_{k}$ = $D / h$ represents the dimension for queries, keys, and values of the Attention mechanism.
Like those in prior studies~\cite{vaswani:nips17:attn,dosovitskiy:iclr21:vit,rombach:cvpr22:sdm,fudong:iccv23:mmst_vit},
$\bm{W}^{Q}_{i} \in \mathbb{R}^{D \times d_{k}}, \bm{W}^{K}_{i} \in \mathbb{R}^{D \times d_{k}}, \bm{W}^{V}_{i} \in \mathbb{R}^{D \times d_{k}}$, and $\bm{W}^{O} \in \mathbb{R}^{D \times D}$
are four learnable projection matrices.

\subsection{Our Sequence Score} 
\label{sec:seq-cam}

Interpreting the biological intelligence encoded in Protein Transformers (PT) 
demands an XAI technique capable of handling long protein sequences 
composed of a substantial number of amino acids.
Previous Transformer-specific XAI methods
\cite{Abnar:acl20:rollout,Voita:acl19:vit-method,Xie:ijcai23:vit-method,Chefer:cvpr21:vit-method,Chefer:iccv21:vit-method} 
have proven effective in elucidating the decision-making processes of various Transformer models 
\cite{dosovitskiy:iclr21:vit,liu:iccv21:swin-vit}.
However, their use is predominantly limited to analyzing shorter token sequences,
due to their substantial computational overhead involved in aggregating attention weights 
across all layers of Transformer Encoders.
This limitation is particularly problematic for interpreting Protein Transformers (PTs), 
which analyze amino acid sequences that often feature extensive lengths, 
thus obstructing their potential to unlock the decision-making processes within PTs.
To address this challenge, 
we introduce the Sequence Score, 
an innovative XAI method 
characterized by its time complexity \textit{growing linearly} with the protein sequence. 
This feature renders it exceptionally suitable for 
decoding the biological intelligence usually embedded within protein sequences.

Given a decision of interest (\eg\ the protease class), 
our Sequence Score,  with respect to the gradients of a well-trained PT model, 
can generate a sequence of important scores, 
based on the primary structure of proteins.
Next, we introduce the details of our Sequence Score technique.
Consider a decision interest of class $c \in [C]$, our Sequence Score technique first calculates the gradient of the logit for the class $c$, 
with respect to feature maps $\bm{A} \in \mathbb{R}^{P \times D}$ of any Transformer block
(\eg\ the last block of Transformer Encoders), 
\ie\ $\frac{\partial y^{c}}{\partial \bm{A}}$.
Here, the term ``logit'' refers to the classification score
before being passed through a sigmoid (or softmax) function 
to produce a probability distribution over the classes.
Then, the neuron importance weight $\bm{w}^{c} \in \mathbb{R}^{D}$, 
is obtained by performing global average pooling over the sequence length (indexed by $j$), \ie\ 
\begin{equation} \label{eq:weight}
    \begin{gathered}
        \bm{w}^{c} = \frac{1}{P} \sum_{j=1}^{P} \frac{\partial y^{c}}{\partial \bm{A}_{j}}. 
    \end{gathered}
\end{equation}
Next, we arrive at the importance score for the $j$-th amino acid $S_{j}^{c}$
by summing up a weighted combination of the feature map activations $\bm{A}$, \ie\
\begin{equation} \label{eq:score}
    \begin{gathered}
        S_{j}^{c} = \sum_{k=1}^{D} \bm{w}_{k}^{c} \bm{A}^{k},  
        \quad  ~j=1,2, \cdots, P. 
    \end{gathered}
\end{equation}
Similar to prior XAI techniques~\cite{selvaraju:iccv17:grad_cam,Chattopadhyay:wacv18:gradcam++},
attention is paid solely to features that positively affect the prediction of interest.
In other words, the negative importance scores should be dropped.
Meanwhile, our preliminary experimental results indicate that if the primary structure of proteins is too long, 
the importance score for each amino acid will be small (\ie\ $ < 0.001$). 
We develop a novel trick to address the two issues simultaneously, 
expressed as follows,
\begin{equation} \label{eq:norm}
    \begin{gathered}
        S_{j}^{c} = \frac{\max (0, ~S_{j}^{c})}{\max(\bm{S}^{c})},
        \quad  ~j=1,2, \cdots, P.
    \end{gathered}
\end{equation} 
Here, the numerator and the denominator of \eqref{eq:norm} serve for dropping the negative scores
and normalizing the positive scores, respectively.
As such, our Sequence Score technique can interpret biological intelligence underlined in PT models 
by revealing their decision-making processes when predicting the functional properties of proteins.
It is noteworthy that the computation of our Sequence Score technique achieves linear time complexity, 
denoted as $\mathcal{O} (D \cdot P)$,
where $D$ represents the hidden dimension of the Transformer Encoders 
and $P$ denotes the length of the protein sequences. 
This efficiency underscores its suitability for analyzing 
the intricate biological intelligence embedded in protein structures.

\section{Experiments and Results}
\label{sec:exp}

\subsection{Experimental Settings}
\label{sec:exp-setup}

\par\smallskip\noindent
{\bf Datasets.}
We conduct experiments across three benchmarks:
\romsm{1}) \textbf{Protein-FN}, having $9,014$ protein data
with their 1D amino acid sequences, 3D protein structures, 
and functional properties;
\romsm{2}) \textbf{Antibiotic Resistance (AR)}~\cite{mcarthur2013comprehensive}, 
containing $3,416$ protein samples,
each associated with its antibiotic type;
and \romsm{3}) \textbf{Metal Ion Binding (MIB)}~\cite{hu:nips22:exploring}, 
offering $7,332$ single protein sequences, 
collected from PDB with annotation as metal ion binding.

%


    
\begin{table*} [!t] 
    \scriptsize
    \centering
    \setlength\tabcolsep{10 pt}
    \caption{
        Model variants of our Sequence Protein Transformers (SPT), 
        with their model details listed below
        }
        \begin{tabular}{@{}cccccc@{}}
            \toprule
            Model        & Layers & Hidden Size $D$ & Head & MLP Size & Parameters \\ \midrule
            SPT-Tiny  & 12    & 192         & 4    & 768      & 5.4M   \\ 
            SPT-Small & 12    & 384         & 6    & 1536     & 21.5M  \\ 
            SPT-Base  & 12    & 768         & 12   & 3072     & 85.5M  \\ \bottomrule
            \end{tabular}
    \label{tab:exp-model-size}
\end{table*}

\par\smallskip\noindent
{\bf Model Variants.}
We set our Sequence Protein Transformers (SPT) configurations 
based on the Transformer settings reported in previous studies
~\cite{dosovitskiy:iclr21:vit,touvron:icml21:deit}.
Three model variants, \ie\ SPT-Tiny, SPT-Small, and SPT-Base, are developed, 
tailored for protein function predictions across different scales of data.
Table \ref{tab:exp-model-size} presents the model details of those SPT variants.
Specifically, all SPT variants are composed of $12$ layers of Transformer blocks,
with their hidden sizes set to $192$, $384$, and $768$,
and their numbers of heads set to $4$, $6$, and $12$,
respectively for the SPT-Tiny, the SPT-Small, and the SPT-Base models.
The MLP sizes are fixed to four times of their corresponding hidden sizes.


\par\smallskip\noindent
{\bf Compared Approaches.} 
As our SPT models belong to the single sequence-based Protein Transformers,
we consider three prominent single sequence-based Protein Language Models (PLMs), \ie\
\textbf{TAPE}~\cite{rao:nips19:tape},
\textbf{ESM-1b}~\cite{rives:pans21:ESM-1b},
and \textbf{ESM-1v}~\cite{meier:nips21:esm_1v},
for baseline comparison.
The hyperparameters for PLM counterparts, if not specified, are set as reported in their original literature.

\par\smallskip\noindent
{\bf Hyperparamters.} 
In sharp contrast to prior PLMs
\cite{rao:nips19:tape,rives:pans21:ESM-1b,meier:nips21:esm_1v},
our SPT models do not require computationally extensive self-supervised pre-training.
Instead, they are all trained from scratch by employing the AdamW~\cite{loshchilov:iclr19:adamw} optimizer
with $\beta_{1} = 0.9$, $\beta_{2} = 0.999$, and a weight decay of $0.05$.
The training epochs for SPT variants are set to $100$, 
including $5$ warmup epochs.
We utilize the cosine decay learning rate schedule~\cite{loshchilov:iclr2016:sgdr}, 
with a base learning rate of $1e-3$ 
and a layer-wise learning rate decay~\cite{clark:iclr19:layer_wise_decay} of $0.75$.
Following~\cite{kaiming:cvpr22:mae}, 
we also apply the label smoothing~\cite{szegedy:cvpr16:label_smoothing}
and 
the path dropping~\cite{huang:eccv16:drop_path},
with their values both set to $0.1$.
As such, our SPT models are superb computation-efficient.
All experiments were conducted on a lab computer with an RTX 4090 GPU,
having its memory usage consistently $\leq 9.6 \%$.


%

\subsection{Comparisons to Protein Language Models} 
\label{sec:exp-overall-comparison}

\par\smallskip\noindent
{\bf Experiments on the Protein-FN Dataset.} 
We conducted experiments on the \textit{Protein-FN} dataset
to evaluate the performance of our SPT models.
Three \sota\ PLMs
mentioned in Section~\ref{sec:exp-setup}
are taken into account as baselines for comparison.
Table~\ref{tab:exp-overall} presents experimental results,
where two notations, \ie\ ``pre-trained'' and ``scratch'',
respectively indicate the counterparts with and without the self-supervised pre-training\footnote{
ESM-1v (scratch) is the same as ESM-1b (scratch)
as they use the same structure but are pre-trained on different datasets.}.
Here, we consider both computational complexity and model performance.
The former is measured by the number of parameters 
and GFLOPs\footnote{GFLOPs, or Giga Floating Point Operations, 
is a metric that quantifies a model's computational complexity. It indicates the number of billion floating-point operations needed by a model per second.
}
calculated under a sequence of amino acids,
while the latter is characterized by 
the top-1 error rates on the training and the test sets.

\begin{table}   [!t] 
    \scriptsize
    \centering
    \setlength\tabcolsep{8 pt}
    \caption{
        Comparisons to \sota\ PLM counterparts on the Protein-FN dataset, 
        where the last two columns respectively report the error rates 
        on the training and the test sets, 
        with the best results shown in bold
        }
        \begin{tabular}{@{}ccccc@{}}
            \toprule
            Methods              & Parameters             & GFLOPs                & Traning Error (\%) & Test Error (\%) \\ \midrule
            TAPE (scratch)       & \multirow{2}{*}{92.4M} & \multirow{2}{*}{21.4} & 2.39                    & 11.59                \\ 
            TAPE (pre-trained)   &                        &                       & 0.34                    & 0.55                 \\ \midrule
            ESM-1b (scratch)     & \multirow{2}{*}{650M}  & \multirow{2}{*}{160}  & 1.11                    & 11.73                \\ 
            ESM-1b (pre-trained) &                        &                       & 1.33                    & 1.86                 \\ \midrule
            ESM-1v (pre-trained) & 650M                   & 160                   & 0.55                    & 0.58                 \\ \midrule
            SPT-Tiny          & \textbf{5.4M}                   & \textbf{1.4}                   & 0.39                    & 0.41                 \\ 
            SPT-Small         & 21.5M                  & 5.1                   & 0.22                    & 0.38                 \\ 
            SPT-Base          & 85.5M                  & 19.4                  & \textbf{0.11}                    & \textbf{0.31}                 \\ \bottomrule
            \end{tabular}
    \label{tab:exp-overall}
\end{table}

We have three observations.
First, our SPT-Tiny model (containing only $5.4$M parameters)
achieves an exceptionally low test set error rate of $0.41\%$,
outperforming all competitors, 
in terms of both computational efficiency and prediction accuracy.
Moreover, its impressive capabilities in protein function predictions 
are achieved with the GFLOPs value of just $1.4$. 
This represents a computational demand at least $15.2 \times$ lower than that of its competitors, 
underscoring the model's remarkable balance between efficiency and accuracy.
Second, our SPT-Base model achieves the best error rate of $0.11\%$ and $0.31\%$ 
respectively on the training and the test sets.
This notable performance underscores a key finding: scaling up our model's size corresponds to a marked enhancement in its capabilities. 
Third, training PLMs from scratch may lead to substantial overfitting.
This issue is evident in the significant discrepancies observed between training and test set performance outcomes,
\eg\ the error rate of $2.39\%$ \textit{v.s.} $11.59\%$ for the TAPE (scratch) model
and of $1.11\%$ \textit{v.s.} $11.73 \%$ for the ESM-1b (scratch) model.
In sharp contrast, our SPT models, despite trained from scratch, show excellent generalization abilities.
This can be attributed to our SPT's design in the amino acid embedding,
which lifts the requirement of self-supervised pre-training 
to learn meaningful protein representations.

\begin{table}   [!t] 
    \scriptsize
    \centering
    \setlength\tabcolsep{8 pt}
    \caption{
        Overall comparisons to PLMs on the AR and MIB datasets,
        with best results shown in bold
        }
        \begin{tabular}{@{}ccccc@{}}
            \toprule
            \multirow{2}{*}{Methods} & \multicolumn{2}{c}{AR}                      & \multicolumn{2}{c}{MIB}                                                                                     \\ \cmidrule(l){2-3} \cmidrule(l){4-5} 
                                     & \multicolumn{1}{c}{GFLOPs}                   & Test Error (\%)                       & \multicolumn{1}{c}{GFLOPs}                            & Test Error (\%) \\ \midrule
            TAPE (scratch)           & \multicolumn{1}{c}{\multirow{2}{*}{106.3}}   & 11.8                                      & \multicolumn{1}{c}{\multirow{2}{*}{65.7}}         & 39.8                \\ 
            TAPE (pre-trained)       & \multicolumn{1}{c}{}                         & 7.3                                      & \multicolumn{1}{c}{}                               & 34.2                \\ \midrule
            ESM-1b (scratch)         & \multicolumn{1}{c}{\multirow{2}{*}{172.1}}   & 11.1                                      & \multicolumn{1}{c}{\multirow{2}{*}{79.1}}         & 40.1                \\ 
            ESM-1b (pre-trained)     & \multicolumn{1}{c}{}                         & 8.6                                      & \multicolumn{1}{c}{}                               & 35.9                \\ \midrule
            ESM-1v (pre-trained)     & \multicolumn{1}{c}{172.1}                    & 9.8                                      & \multicolumn{1}{c}{79.1}                           & 33.3                \\ \midrule
            SPT-Tiny                 & \multicolumn{1}{c}{\textbf{19.2}}                     & 5.7                                  & \multicolumn{1}{c}{\textbf{5.6}}              & 37.1            \\ 
            SPT-Small                & \multicolumn{1}{c}{31.2}                     & 4.5                                  & \multicolumn{1}{c}{18.5}                               & 32.7            \\ 
            SPT-Base                 & \multicolumn{1}{c}{105.6}                    & \textbf{4.3}                                  & \multicolumn{1}{c}{65.7}                      & \textbf{32.1}            \\ \bottomrule
        \end{tabular}
        \label{tab:overall-datasets}
\end{table}

\par\smallskip\noindent
{\bf Experiments on the AR and MIB Datasets.} 
Here, we conducted experiments on the two widely-used benchmarks,
\ie\ Antibiotic Resistance (AR) and Metal Ion Binding (MIB),
for further exhibiting the effectiveness of our SPT models for protein function predictions.
Table~\ref{tab:overall-datasets} presents experimental results.
It is observed that our SPT-Base model beats all PLM counterparts on both datasets, 
with the best test error rates of $4.3\%$ and $32.1 \%$ on the AR and MIB datasets.
In addition, our SPT models stand out for their computational efficiency. 
Taking the
 the AR dataset for instance,  
our SPT-Tiny model not only achieves a low test error rate of $5.7\%$ 
but did so with just $19.2$ GFLOPs of computational demand. 
This efficiency makes it at least $5.5$ times faster than previous PLMs, 
marking a significant advancement in processing speed and energy consumption.
Finally, without self-supervised pre-training,
previous PLMs suffer from a significant performance degradation
(See $3$rd \textit{v.s.} $4$th rows and $5$th \textit{v.s.} $6$th rows).
On the contrary, our SPT models, though trained from scratch,
achieves superb prediction performance outcomes.
This can be attributed to the effectiveness of our novel protein embedding mechanism.

We've further evaluated our SPT models, 
comparing them with traditional bioinformatics approaches 
and conducting detailed ablation studies,
with corresponding experimental results deferred to 
Appendices A.1 and A.2, respectively.

\subsection{Evaluation on Our Sequence Score Technique} 
\label{sec:exp-eval-seq-cam}

We consider two metrics proposed in the prior study~\cite{alvarez:nips18:xai_metric},
\ie\ \textbf{Faithfulness} and \textbf{Stability},
to evaluate the efficacy of our Sequence Score technique for explaining Protein Transformers.
In this section, we present the experimental results regarding the \textit{faithfulness} metric. 
The evaluation of the \textit{stability} of our Sequence Score technique is detailed in 
Appendix B.1 of the supplementary materials 
for conserving space.
In particular, \textit{faithfulness} measures the degree to which the importance values
which are attributed to amino acids,
are aligned with their actual impact on the final prediction,
expecting that amino acids with substantial effects will receive correspondingly high importance scores.

%
Here, we adopt the \textit{deletion} method~\cite{petsiuk:bmvc18:deletion} 
to evaluate the \textit{faithfulness} of our Sequence Score technique. 
Its key idea is to observe accuracy degradation incurred by masking a certain ratio (or number) of amino acids,
with masked amino acids chosen based on their importance scores,
\ie\ those with the highest scores \textit{v.s.} those with the lowest scores.
A larger drop in prediction accuracy, 
resulted from masking amino acids with high importance scores than with low scores,
indicates that the assigned scores are well aligned with amino acids' actual significance to the final predictions.

\begin{figure} [!t] 
   \centering
    \captionsetup[subfigure]{justification=centering}
    \begin{subfigure}[t]{0.35\textwidth}
        \centering
        \includegraphics[width=\textwidth]{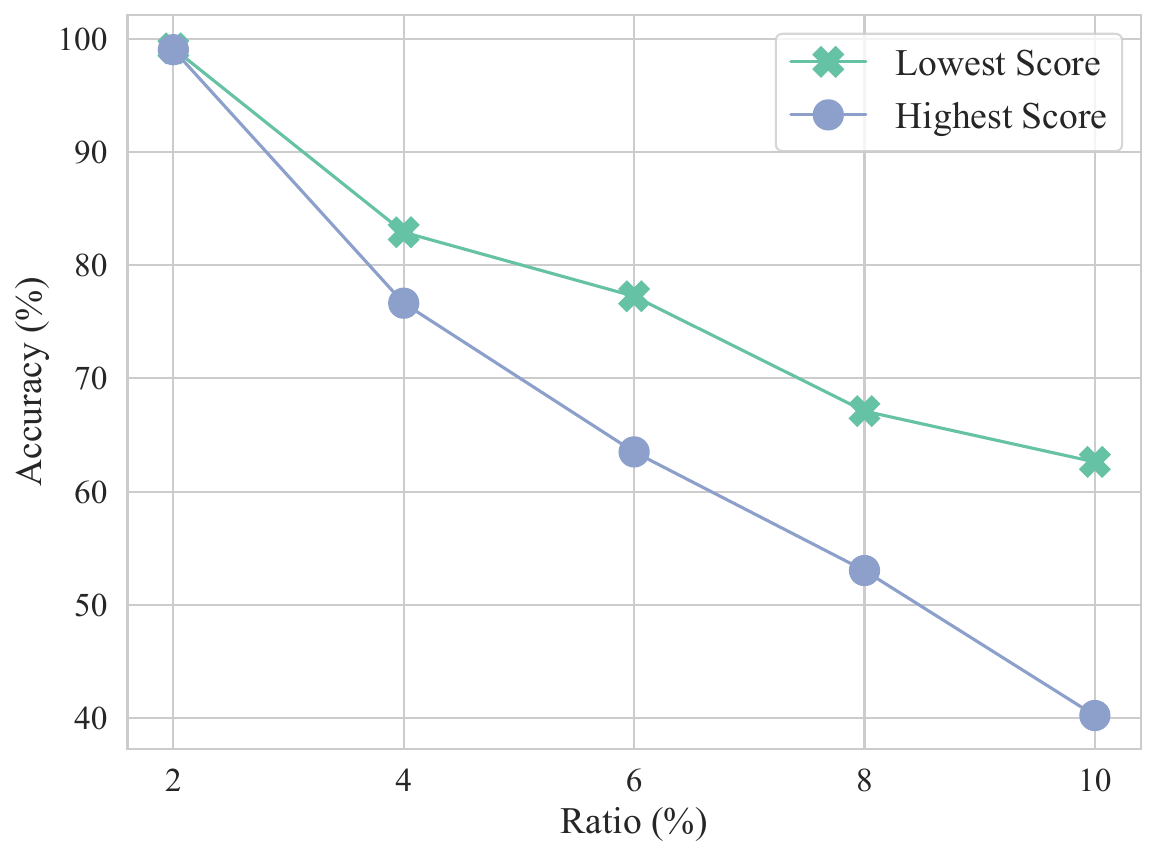}
        \caption{Ratio}
        \label{fig:exp-val-cam-ratio-masking}
    \end{subfigure}
    \begin{subfigure}[t]{0.35\textwidth}
        \centering
        \includegraphics[width=\textwidth]{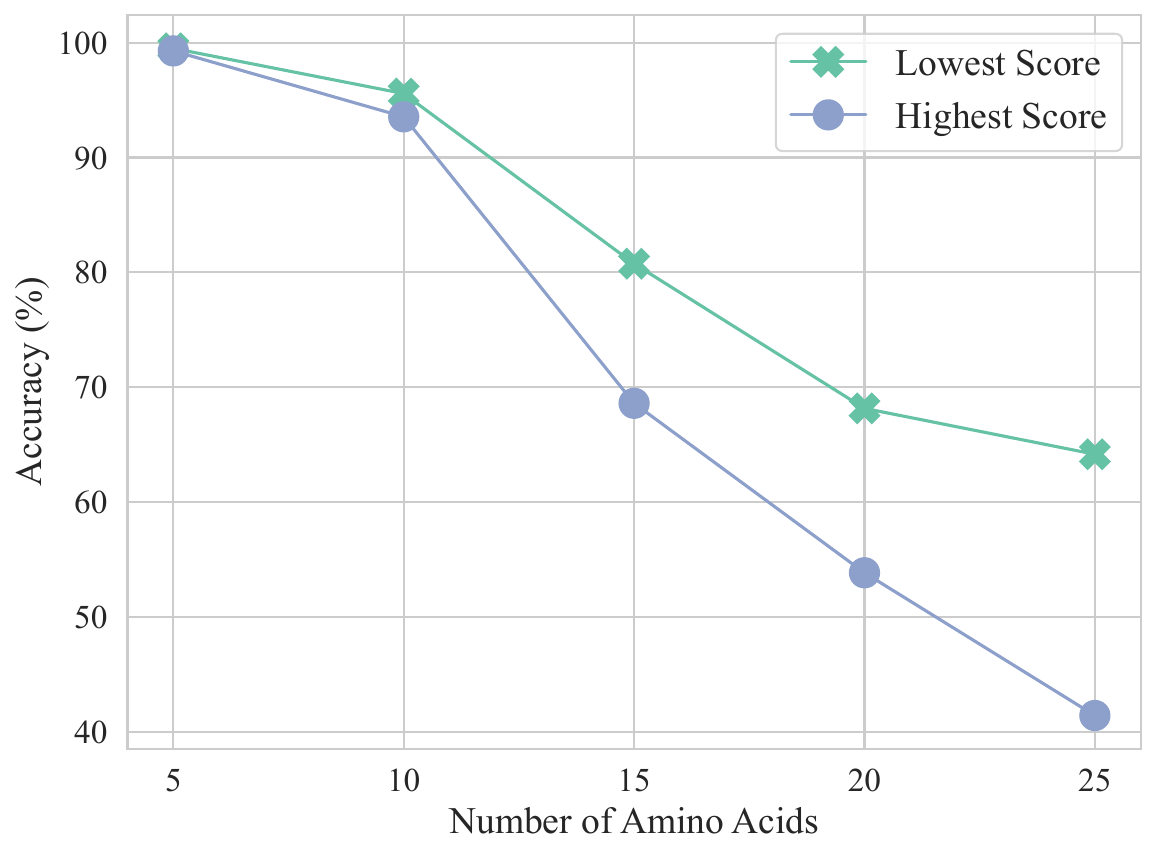}
        \caption{Number}
        \label{fig:exp-val-cam-num-masking}
    \end{subfigure}
    \caption{
        Comparisons of prediction performance by masking amino acids with the highest and the lowest importance scores under
        (a) masking a certain ratio of amino acids and (b) masking a specific number of amino acids.
    }
    \label{fig:exp-val-cam-masking}
\end{figure}

Figure~\ref{fig:exp-val-cam-ratio-masking} illustrates the results of our experiments 
that involve selectively masking amino acids at various ratios.
Our findings reveal a consistent pattern: 
masking amino acids identified as highly important invariably leads to
a larger decline in prediction performance,
versus masking amino acids with lower importance scores.
For example, when the masking ratio is set to $10\%$, 
masking amino acids that have the highest importance scores results in performance degradation 
to be $22.41 \%$ greater than that resulted from masking those with the lowest importance scores.
Additionally, lifting the masking ratio from $2\%$ to $10\%$ yields a noticeably faster decline in prediction performance 
when masking amino acids with the highest scores (see the blue line) than with the lowest scores (see the green line).
This demonstrates a direct relationship between the importance scores assigned to the amino acids and their actual influence levels on the predictive accuracy of the model.

Complementing these findings, 
Figure~\ref{fig:exp-val-cam-num-masking} illustrates similar trends under a different experimental setting, 
where various numbers of amino acids are masked.
As the number of masked amino acids increases from $5$ to $25$,
prediction performance is observed to degrade significantly faster 
when masking amino acids with the highest scores than with the lowest scores.
Specifically, a masking number of $25$ leads to a substantial larger performance decline, \ie\ by $58.16\%$, when 
masking amino acids with the highest importance scores than with the lowest scores, 
\ie\ only by $35.41\%$.
These statistical observations confirm that 
our Sequence Score technique adheres to the principle of \textit{faithfulness} 
when interpreting Protein Transformers.

We have also conducted experiments to assess the \textit{faithfulness} of our Sequence Score technique
by simulating protein mutations,
with their results deferred to 
Appendix B.2 of supplementary materials.

\subsection{Discovery of Catalytic Triad in Serine Proteases} 
\label{sec:exp-cata-triad}

\begin{figure}  [!t] 
    \centering
    \centering
    \includegraphics[width=0.55\textwidth]{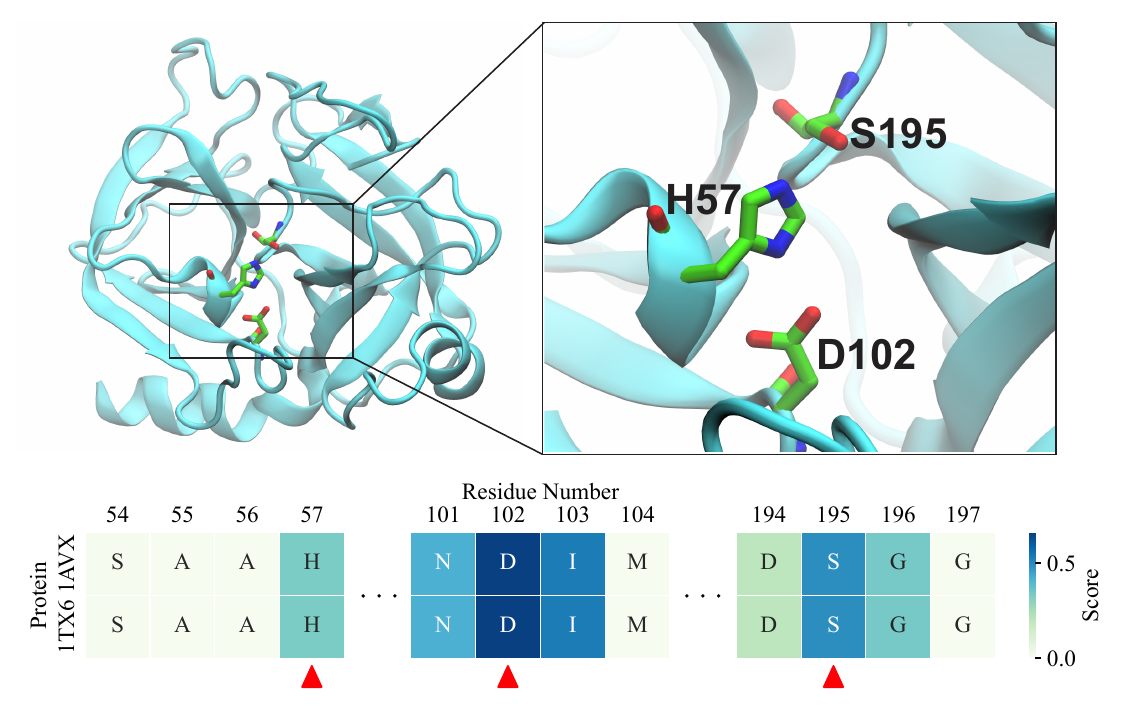}
    \caption{
        Illustration of the catalytic triad,
        \ie\ ``His57-Asp102-Ser195'',
        in the two proteins of serine proteases.
        Here, ``His'', ``Asp'', and ``Ser'' are abbreviated as ``H'', ``D'' and ``S'', respectively.
        The figure in the upper section utilizes the protein ``1TX6'' as an example to show the 3D structure of the catalytic triad,
        while the heatmap in the lower section visualizes the importance scores
        generated by our Sequence Score technique,
        with amino acids corresponding to 
        the catalytic triad distinctly marked by red triangles.
    }
    \label{fig:exp-pattern-cata-triad}
\end{figure}

This section further interprets biological intelligence 
resided in Protein Transformers
by unveiling its discovery of the catalytic triad in serine proteases.
Serine proteases, a group of proteases, are crucial enzymes involved in a myriad of biological functions, including digestion, immune response, and blood coagulation. 
At the heart of their catalytic mechanism lies the catalytic triad,
a set of three coordinated amino acids, 
usually following the pattern of ``His57-Asp102-Ser195''.
This triad forms a potent synergistic unit essential for the enzyme's function. 
Figure~\ref{fig:exp-pattern-cata-triad} depicts importance scores for the two proteins of serine proteases.
It is obvious that our SPT models can identify the significance of the catalytic triad
(marked by red triangles) for serine proteases.
These results further confirm that our Protein Transformers 
can capture biological intelligence inherent within protein sequences.

It is worth noticing that the catalytic triad's importance extends beyond its biochemical role; 
its evolutionary conservation across various serine proteases underscores a fundamental mechanism critical to many physiological processes. 
Moreover, the detailed understanding of this triad, including the specific residue numbers, 
has been instrumental in the design of targeted pharmaceutical inhibitors 
to modulate serine protease activity in treating various diseases, 
such as cancer, inflammatory disorders, and coagulopathies.
Hopefully, the capability of our SPT models to discover the catalytic triad of serine proteases
can shed light on future studies, 
aided by the Protein Transformers for understanding the biochemical, physiological, and pharmaceutical processes.

\section{Conclusion}
\label{sec:conslusion}

This work has explored the capabilities of Protein Transformers
in capturing biological intelligence resided in protein sequences.
To achieve our goal,
we first introduced the \textit{Protein-FN} dataset,
offering over $9000$ protein sequence data as well as their functional properties
created laboriously by biological experts.
Then, we developed the Sequence Protein Transformers (SPT), 
a computationally efficient Transformer architecture,
able to precisely predict the functional properties of proteins
by leveraging their primary structures. 
Thanks to its novel protein embedding mechanism,
our SPT models can achieve superb prediction performance 
without the requirement of self-supervised pre-training.
Finally, we have developed the Sequence Score, 
a novel Explainable Artificial Intelligence (XAI) technique 
that advances beyond current Transformer-specific XAI solutions in terms of computational efficiency. 
This efficiency increases linearly with the length of protein sequences, 
making it well-suited for analyzing the complex biological intelligence encoded within Protein Transformers.
Extensive experimental results exhibited that 
our SPT models are efficient and effective in predicting the functional properties of proteins.
Moreover, the devised Sequence Score technique helps reveal 
that our SPT models can capture important patterns underlying protein sequences,
with these patterns aligning closely with the domain knowledge in the biology community.
This demonstrates the capabilities of our Protein Transformers in capturing biological intelligence resided in protein sequences.

\section*{Acknowledgments}
This work was supported in part by NSF under Grants  2019511 and 2425812.  Any opinions and findings expressed in the paper are those of the authors and do not necessarily reflect the views of funding agencies.

\bibliographystyle{plain}
\bibliography{main}

\newpage
\appendix

\section*{Outline}
This document supplements the main paper in two aspects.
%
%
First, Section~\ref{sec:sup-exp-spt} presents supporting experimental results for our SPT models.
Second, Section~\ref{sec:sup-exp-seq-cam} provides additional evaluations of our Sequence Score technique.

\section{Supporting Experiments on Our SPT Models} 
\label{sec:sup-exp-spt}

\subsection{Comparisons to Bioinformatics Approaches} 
\label{sec:sup-exp-seq-vit}

This section supports the main paper by conducting experiments on our \textit{Protein-FN} dataset to 
evaluate the performance of our SPT models against 
two conventional bioinformatics algorithms,
\ie\ MUSCLE~\cite{edgar2004muscle} and TM-aligh~\cite{zhang2005tm_align},
in predicting the functional properties of proteins.
Table~\ref{tab:exp-comp-bio} lists the experimental results.
We have two observations.
First, our smallest model, SPT-Tiny, 
demonstrates superior accuracy in protein function prediction,
surpassing the error rates of MUSCLE and TM-align by $4.6 \%$ and $0.2 \%$, respectively.
Second, progressing from SPT-Tiny to SPT-Base, we observe a reduction in the error rate from $0.41 \%$ to $0.31 \%$. This trend indicates that scaling up the size of our SPT models correlates with increased predictive precision, underscoring the enhanced capabilities of larger SPT models in accurately predicting protein functions.
The empirical findings presented here, together with those detailed in 
Section 5.2 
of the main paper, demonstrate that our SPT models excel in predicting the functional properties of proteins, 
outperforming both traditional bioinformatics methods and modern Protein Language Models (PLMs).

\begin{table}    [!t] 
    \scriptsize
    \centering
    \setlength\tabcolsep{10 pt}
    \caption{
        Comparisons to conventional bioinformatics approaches in protein function predictions, where the error rate on the test set is reported, with the best result shown in bold
        }
    \begin{tabular}{@{}cc@{}}
    \toprule
    Methods      & Error Rate ($\%$) \\ \midrule
    MUSCLE       & 5.05            \\
    TM-align     & 0.61            \\ \midrule
    SPT-Tiny  & 0.41            \\
    SPT-Small & 0.38            \\
    SPT-Base  & \textbf{0.31}            \\ \bottomrule
    \end{tabular}
    \label{tab:exp-comp-bio}
\end{table}

\subsection{Ablation Studies on Our SPT Model} 
\label{sec:exp-as}

\par
\begin{figure}    [!t]  
    \centering
     \begin{subfigure}[t]{0.35\textwidth}
        \centering
        \includegraphics[width=\textwidth]{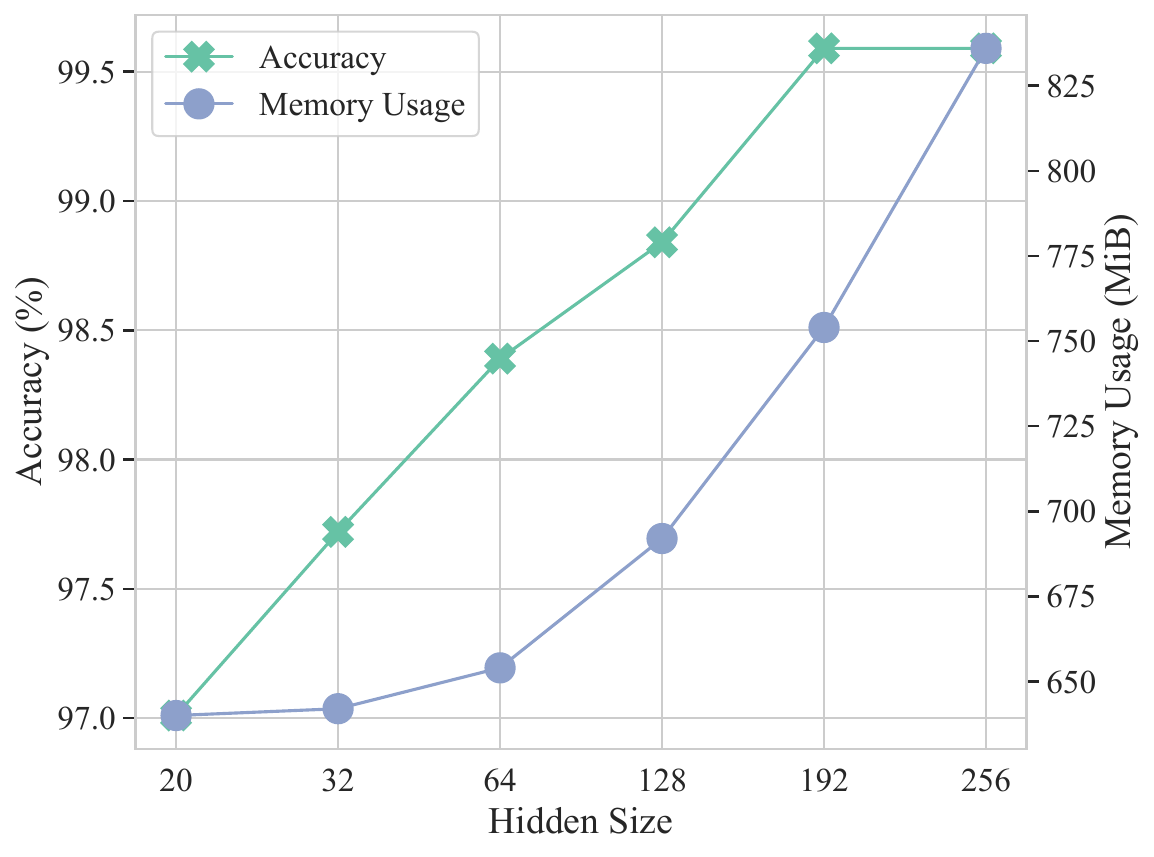}
        \caption{Hidden size}
        \label{exp-as-embed}
    \end{subfigure}
    \begin{subfigure}[t]{0.35\textwidth}
        \centering
        \includegraphics[width=\textwidth]{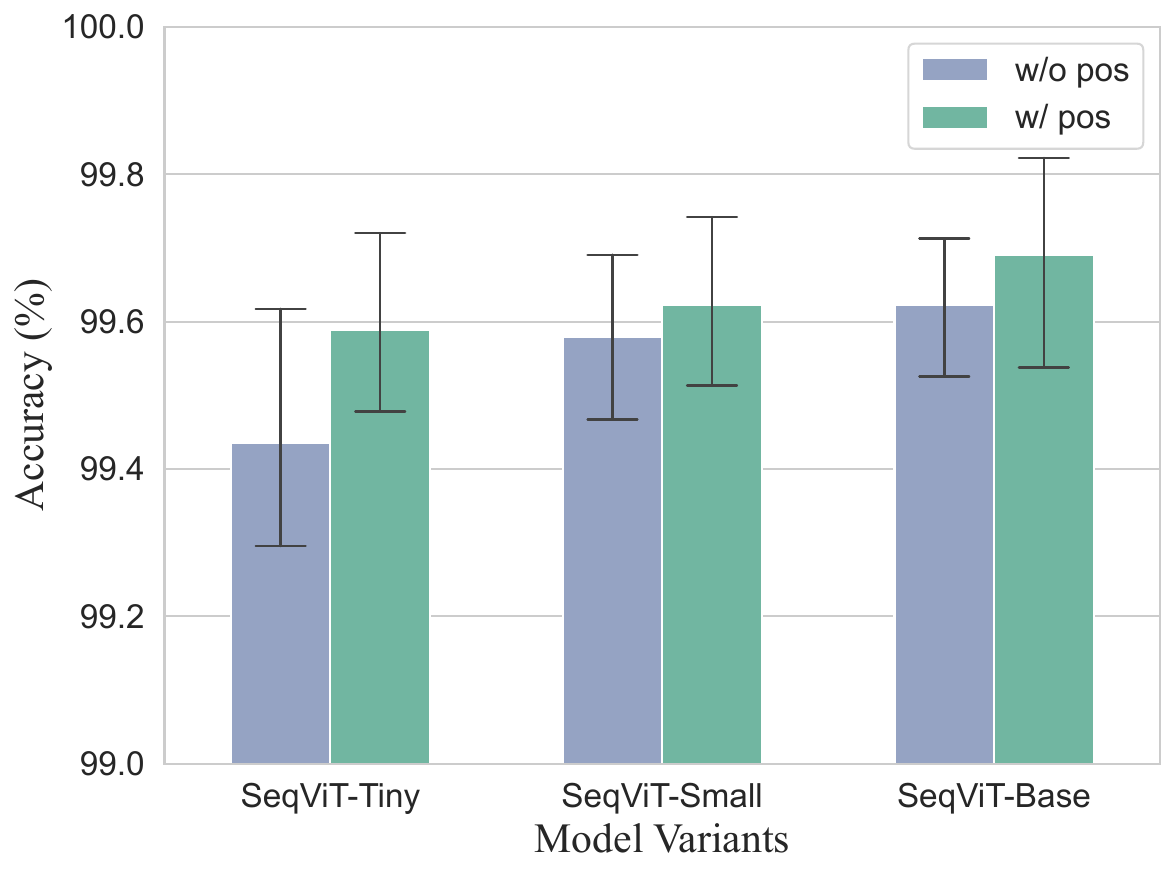}
        \caption{Positional embedding}
        \label{exp-as-pos}
    \end{subfigure}
    \caption{
        Illustration of how (a) the hidden size and (b) the positional embedding affect prediction performance.
    }
    \label{fig:exp-as}
\end{figure}

\smallskip\noindent
{\bf Hidden Size.} 
We have conducted experiments to explore how the hidden size in our SPT-Tiny model 
affects protein function predictions by varying its dimension from $20$ to $256$.
Figure~\ref{exp-as-embed} illustrates the experimental results 
in terms of prediction accuracy and computational overhead.
In this context, the computational overhead is quantified in terms of Mebibytes (MiB), 
which signifies the GPU memory usage during training.
We discover that an increase in the hidden size positively affects prediction performance 
but negatively influences the computational burden.
The best trade-off is achieved at the hidden size equal to $192$, 
where our  model enjoys a superb prediction accuracy of $99.59 \%$ 
at the expense of $754$ MiB memory consumption.
Meanwhile, a larger hidden dimension cannot further improve prediction performance
if the hidden size exceeds $192$.
This indicates that the model's learning ability is bottlenecked by other factors,
\eg\ the data itself.

\par\smallskip\noindent
{\bf Positional Embedding.} 
Unlike fixed start points for conventional vision tasks,
the start point of the protein sequence, indicated by the ``residue number'' in the PDB data,
may vary in real scenarios,  
\eg\ $-7$ in ``6H8R'' \textit{v.s.} $1$ in ``5QEW'' \textit{v.s.} $16$ in ``1A5G''.
%
%
Here, we explore whether the positional embedding can benefit protein function predictions, 
when only the relative positional information is provided.
Figure~\ref{exp-as-pos} shows prediction performance using our SPT models
with or without the positional embedding. 
In all scenarios, our SPT models perform better with the positional embedding (\ie\ the green bars) than without the positional embedding (\ie\ the blue bars),
exhibiting prediction accuracy improvement by $0.15 \%$, $0.04 \%$, and $0.07 \%$,
respectively for the SPT-Tiny, the SPT-Small, and the SPT-Base models.
The empirical results evidence that the positional embedding 
can benefit protein function predictions by learning the relative positional information.

\section{Additional Evaluation on Our Sequence Score Technique} 
\label{sec:sup-exp-seq-cam}

\subsection{The Stability of Our Sequence Score Technique} 
\label{sec:sup-stab}
This section supports the main paper by presenting the evaluation of our Sequence Score technique regarding the metric of \textit{stability} mentioned in the prior study~\cite{alvarez:nips18:xai_metric}.
Specifically, \textit{stability} assesses the consistency of the importance scores assigned to input data,
positing that proteins with analogous primary structures should obtain similar importance scores.

\begin{figure*}  [!t] 
    \centering
    \captionsetup[subfigure]{justification=centering}
    \begin{subfigure}[t]{0.48\textwidth}
        \centering
        \includegraphics[width=\textwidth]{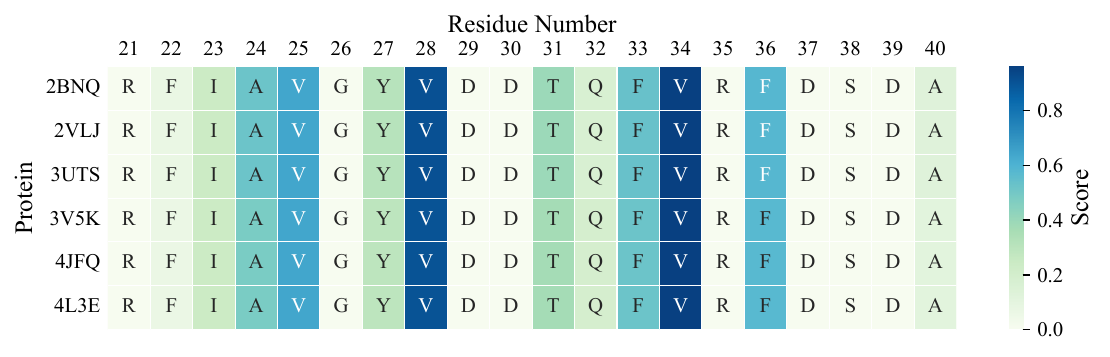}
        \caption{Heatmap for the receptor class}
        \label{exp-stability-receptor}
    \end{subfigure}
    \begin{subfigure}[t]{0.48\textwidth}
        \centering
        \includegraphics[width=\textwidth]{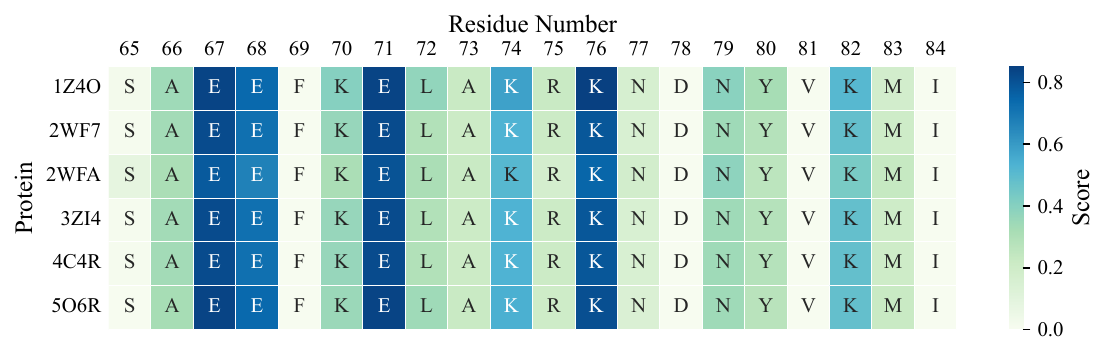}
        \caption{Heatmap for the isomerase class}
        \label{fig:exp-stability-isomerase}
    \end{subfigure}
    \begin{subfigure}[t]{0.48\textwidth}
    \centering
    \includegraphics[width=\textwidth]{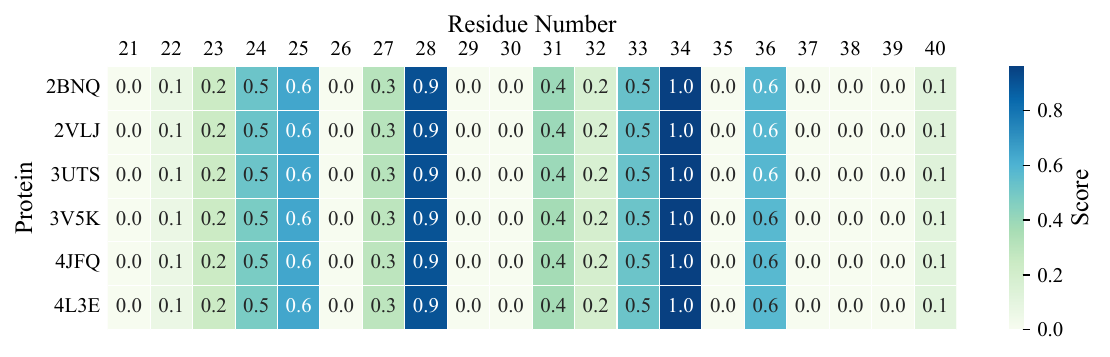}
    \caption{Score values for the receptor class}
    \label{exp-stability-receptor-score}
    \end{subfigure}
    \begin{subfigure}[t]{0.48\textwidth}
    \centering
    \includegraphics[width=\textwidth]{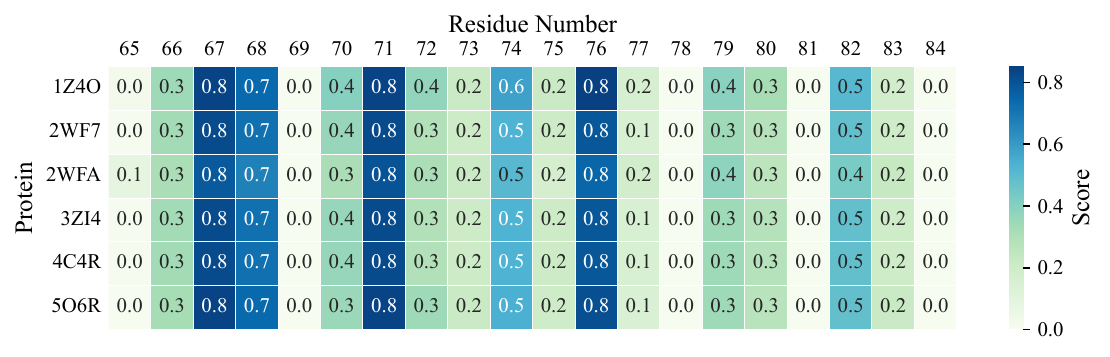}
    \caption{Score values for the isomerase class}
    \label{fig:exp-stability-isomerase-score}
    \end{subfigure}
    \caption{
        The heatmaps of importance scores for (a) the ``receptor'' and (b) the ``isomerase'' classes, with (c) and (d) respectively displaying the 
        corresponding score values for the ``receptor'' and the ``isomerase'' classes.
    }
    \label{fig:exp-stability}
\end{figure*}

We analyze the heatmap of importance scores
to evaluate the \textit{stability} of our Sequence Score technique.
Specifically, proteins with similar primary structures 
are selected intentionally to ascertain the consistency of their corresponding importance scores.
Figures~\ref{exp-stability-receptor} and \ref{fig:exp-stability-isomerase} respectively depict the heatmaps of importance scores for 
the ``receptor" and the ``isomerase'' classes,
with their corresponding score values detailed in Figures~\ref{exp-stability-receptor-score} and \ref{fig:exp-stability-isomerase-score}.
It is obvious that our Sequence Score technique consistently attributes similar scores to structurally comparable entities. 
For example,
no significant score differences are observed in the ``receptor" class, and discrepancies in the ``isomerase'' class are marginal, 
with the largest variance being only $0.1$,
\eg\ protein ``1Z4O'' versus protein ``2WFA'' at residue number $79$.
The statistical evidence validates that our approach reliably upholds the principle of \textit{stability},
thereby effective in interpreting the decision-making processes of Protein Transformers.

\subsection{The Faithfulness of Our Sequence Score Technique} 
\label{sec:sup-faithful}

This section supports the main paper by further evaluating the \textit{faithfulness} of our Sequence Score technique by simulating protein mutations\footnote{Here, we selectively substitute a specific proportion (or number) of amino acids to mimic protein mutations.}.
Its key idea is to observe performance degradation resulting from mutating a certain ratio (or number) of amino acids, selected based on their importance scores. 
This includes amino acids with the highest importance scores versus those with the lowest scores. 
A larger drop in prediction accuracy from mutating high-importance-score amino acids, compared to those with low scores, suggests that the scores effectively reflect the true significance of amino acids in the final predictions.

\begin{figure}   [!t] 
    \captionsetup[subfigure]{justification=centering}
    \centering
    \begin{subfigure}[t]{0.35\textwidth}
        \centering
        \includegraphics[width=\textwidth]{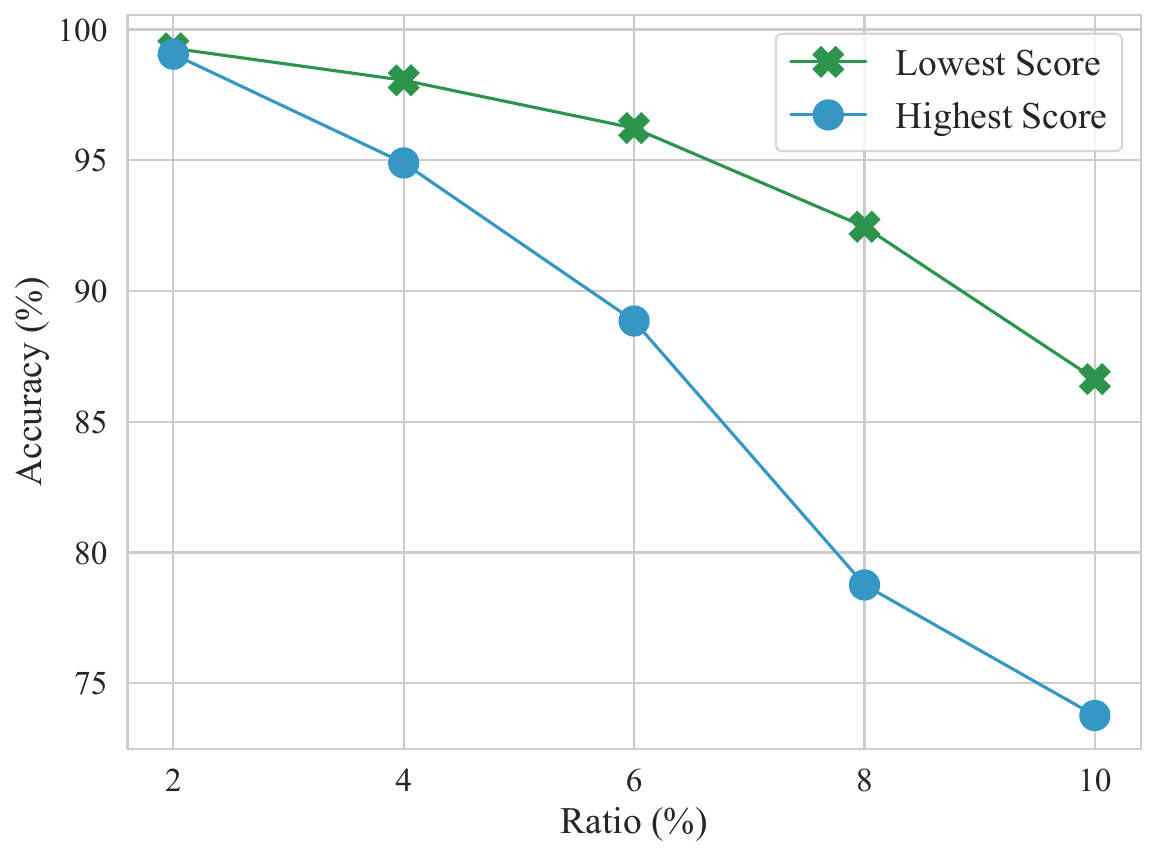}
        \caption{Ratio}
        \label{fig:exp-val-cam-ratio-random}
    \end{subfigure}
    \begin{subfigure}[t]{0.35\textwidth}
        \centering
        \includegraphics[width=\textwidth]{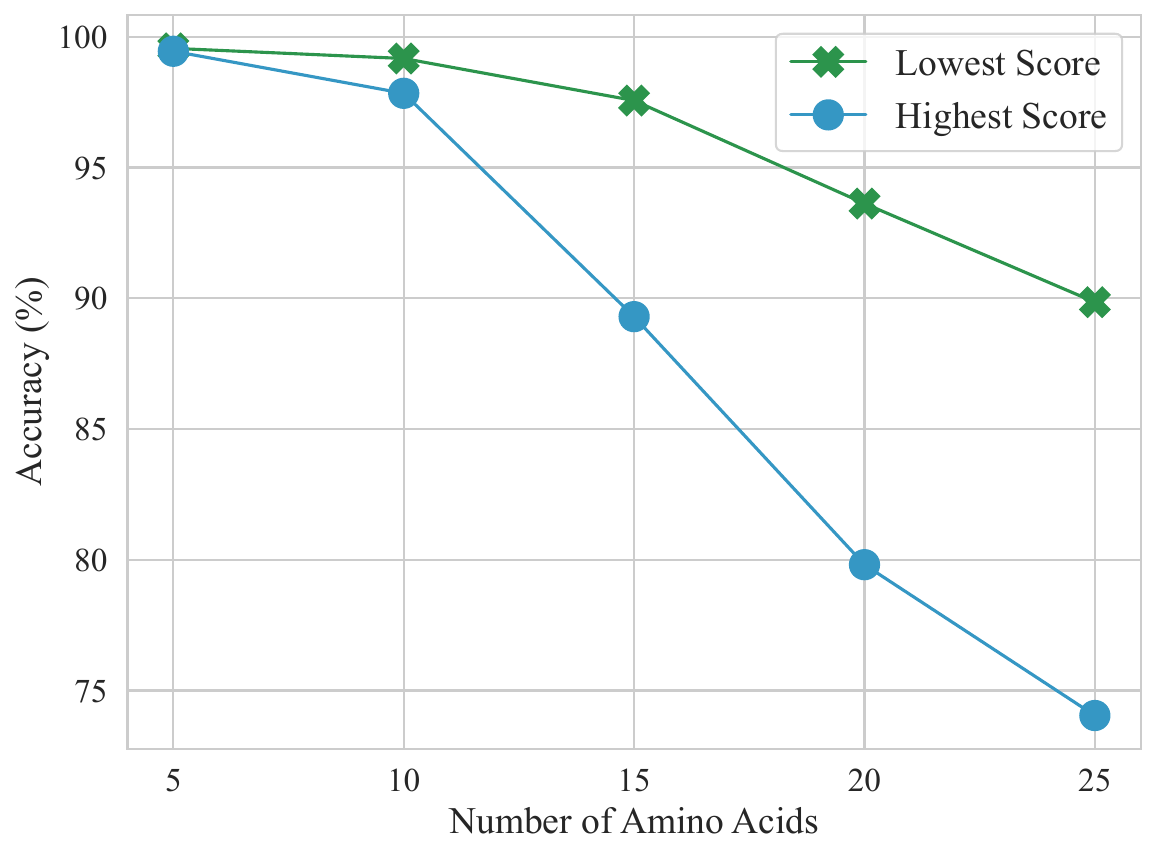}
        \caption{Number}
        \label{fig:exp-val-cam-num-random}
    \end{subfigure}
    \caption{
        Comparisons of prediction performance by simulating protein mutations that occur in amino acids with the highest importance scores and in those with the lowest scores.
        Here, we consider two scenarios: mutations occurring in
        (a) a certain ratio of amino acids and (b) a specific number of amino acids.
    }
    \label{fig:exp-val-cam-random}
\end{figure}

Figure~\ref{fig:exp-val-cam-ratio-random} shows the experimental results that involve the mutations of amino acids at various ratios.
The results reveal that mutations occurring in amino acids recognized as highly important consistently result in a greater decrease in prediction performance, compared to those with lower scores.
For example, when the ratio is set to $10\%$, 
mutations that involve amino acids that have the highest importance scores result in performance degradation 
to be $12.87 \%$ greater than that resulted from those with the lowest importance scores.
Moreover, increasing the mutation ratio from $2\%$ to $10\%$ results in a markedly more rapid decrease in prediction accuracy, when mutations occur in amino acids with the highest importance scores (see the blue line) compared to those with the lowest scores (see the green line).

Figure~\ref{fig:exp-val-cam-num-random} displays the experimental results achieved through mutating different quantities of amino acids.
As the number of mutated amino acids increases from $5$ to $25$,
there is a more rapid and pronounced decline in prediction performance,
when mutations involve amino acids with the highest importance scores, as opposed to those with the lowest scores.
In particular, the mutation of $25$ amino acids leads to a substantial larger performance decline, \ie\ by $25.54\%$, 
when it involves amino acids with the highest importance scores, 
compared to those with the lowest scores, 
\ie\ by $9.73\%$.
These findings corroborate the discovery outlined in our main paper: the Sequence Score technique reliably assigns high-importance scores to amino acids crucial for final predictions. Consequently, our approach effectively adheres to the principle of \textit{faithfulness},
thereby useful for exploring biological intelligence bided in Protein Transformers.

\end{document}